\documentclass[10pt,twocolumn,letterpaper]{article}

\pdfoutput=1
\usepackage{iccv}
\usepackage{times}
\usepackage{epsfig}
\usepackage{graphicx}
\usepackage{amsmath}
\usepackage{amssymb}

\usepackage[breaklinks=true,bookmarks=false]{hyperref}  

\iccvfinalcopy 

\def\iccvPaperID{****} 

\usepackage{booktabs}

\newcommand{\dc}{}

\pdfoutput=1

\begin{document}

\title{\dc BAAI-VANJEE Roadside Dataset: Towards the Connected Automated Vehicle Highway technologies in Challenging Environments of China} 

\author{
Yongqiang Deng$^{1}$, Dengjiang Wang$^{1}$, Gang Cao$^{2}$, \\
Bing Ma$^{1}$, Xijia Guan$^{1}$, Yajun Wang$^{1}$,
Jianchao Liu$^{1}$, Yanming Fang$^{1}$, Juanjuan Li$^{1}$\\
$^{1}$ VanJee Technology, Beijing, China\\
$^{2}$ BeiJing Academy of Artificial Intelligence, Beijing, China\\
{\tt\small \{dengyongqiang, wangdengjiang, mabing, guanxijia\}@wanji.net.cn}\\
{\tt\small \{wangyajun, liujianchao, fangyanming, lijuanjuan\}@wanji.net.cn}\\
{\tt\small caogang@baai.ac.cn}
}

\maketitle

\begin{abstract}

As the roadside perception plays an increasingly significant role in the Connected Automated Vehicle Highway(CAVH) technologies, there are immediate needs of challenging real-world roadside datasets for bench marking and training various computer vision tasks such as 2D/3D object detection and multi-sensor fusion. 
In this paper, we firstly introduce a challenging BAAI-VANJEE roadside dataset which consist of LiDAR data and RGB images collected by VANJEE smart base station placed on the roadside about 4.5m high. This dataset contains 2500 frames of LiDAR data, 5000 frames of RGB images, including 20\% collected at the same time. It also contains 12 classes of objects, 74K 3D object annotations and 105K 2D object annotations. Compared with existing datasets, BAAI-VANJEE features data in varying weather, traffic conditions and high-quality annotation data from God’s perspective. 
By providing a real complex urban intersections and highway scenes, we expect the BAAI-VANJEE roadside dataset will actively assist the academic and industrial circles to accelerate the innovation research and achievement transformation in the field of intelligent transportation in big data era.

\end{abstract}

\section{Introduction}

The main limitations of the vehicle-based technologies are the high concentration of onboard technologies and sensors, which make the large-scale deployment of technologies difficult in the economic and social level.
The CAVH technologies has increasingly recognized~\cite{feng2017ieee}~\cite{li2017icv}.
Different from 2D/3D object detection of the vehicle-based, the 2D/3D object detection based on roadside equipments needs
consideration of the perception problems caused by different sensor installation positions.
Self-driving cars largely exploit supervised machine learning algorithms to perform tasks such as 2D/3D object detection~\cite{yu2019cavh}~\cite{shi2019ieee}~\cite{lang2019ieee}.
Existing vehicle-based multi-sensor processing methods cannot be directly used for roadside multi-sensors, and specific methods for roadside conditions must be additionally developed.
However, most existing datasets have not paid enough attention to these tasks or were not able to fully address them.
The KITTI vision Benchmark Suite~\cite{geiger2013kitti} take advantage of autonomous driving platform Annieway to develop novel challenging real-word computer vision benchmark tasks on stereo, optical flow, visual odometry, 3D object detection and 3D tracking.
ApolloScape~\cite{huang2018apolloscape} advanced open tools and datasets for autonomous driving aiming at 3D perception LiDAR object detection and tracking, including about 100K image frames, 80K LiDAR point cloud and 1000km trajectories for urban traffic.
The Audi Autonomous Driving Dataset(A2D2)~\cite{geyer2020a2d2} consists of simultaneously recorded images and 3D point clouds, together with 3D bounding boxes, semantic segmentation, instance segmentation, and data extracted from automotive bus.
A*3D~\cite{pham2019a3d} has been released with 230K 3D object annotation in 39179 LiDAR point cloud frames and corresponding frontal-facing RGB images. {D$^2$-City} dataset~\cite{che2019d2city} provides more than 1000 videos recorded in 720p HD or 1080p FHD from front-facing dashcams, with detailed annotations for object detection and tracking.

While mentioned datasets above such as A2D2 and A*3D also provide both LiDAR and camera data, more recent datasets have put emphasis on providing autonomous driving datasets collected by self-driving cars~\cite{caesar2020ieee}~\cite{cordts2016ieee}~\cite{yu2018arx}.
For CAVH roadside perception, there are relatively few public roadside datasets, especially 3D point cloud datasets, which cannot meet the current needs for iterative optimization of roadside perception models.

To address this issues, we introduce a new dataset, called BAAI-VANJEE dataset, serving for CAVH. This dataset contains $2500$ frames of LiDAR data and $5000$ frames of RGB images, including 20\% data collected at the same time.
It has 74K human-labeled 3D object annotations and 105K 2D object with detailed annotations captured at different times(day, night) and weathers(sun, cloud, rain).
By providing a real complex urban intersections and highway scenes, we expect the BAAI-VANJEE roadside dataset will actively assist the academic and industrial circles to accelerate the innovation research and achievement transformation in the field of intelligent transportation in big data era.In summary:
\begin{itemize}
  \item We release a new challenging BAAI-VANJEE roadside dataset for CAVH with highly diverse scenes in China.
  \item For all the collected data with $2500$ frames of LiDAR data and $5000$ frames of RGB images, we annotate detailed detection information of traffic participants, including the 2D/3D bounding box coordinates for $12$ classes.
  \item we support three tasks including 2D object detection, 3D object detection and multi-sensor fusion.
  
\end{itemize}

The BAAI-VANJEE dataset is available online at \url{https://data.baai.ac.cn/data-set}.

\section{Dataset}
\begin{figure}[!htbp]
    \begin{center}
        \includegraphics[width=0.815\linewidth]{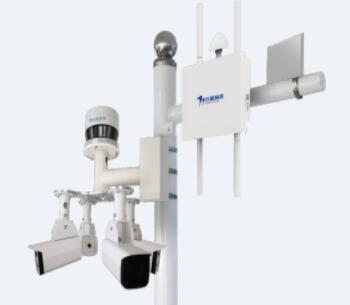} \\
    \end{center}
    \caption{Sensor setup for BAAI-VANJEE data collection platform. The acquisition equipment consists of a rotating VANJEE LiDAR and four network cameras placed under the LiDAR.}
    \vspace{-0.2in}
    \label{fig:long}
\end{figure}
\begin{figure}[!htbp]
    \begin{center}
        \includegraphics[width=0.815\linewidth]{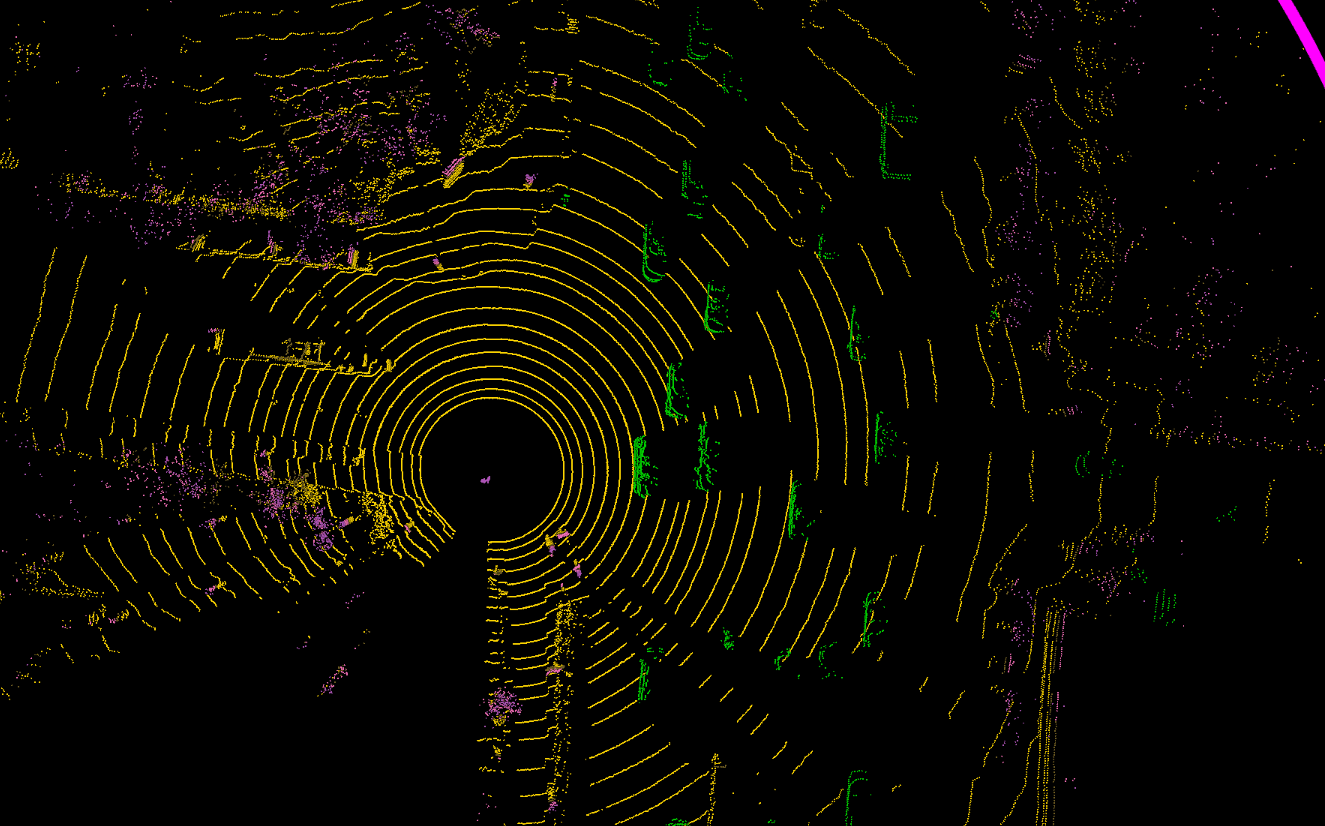} \\ \vspace{0.05in}
        \includegraphics[width=0.815\linewidth]{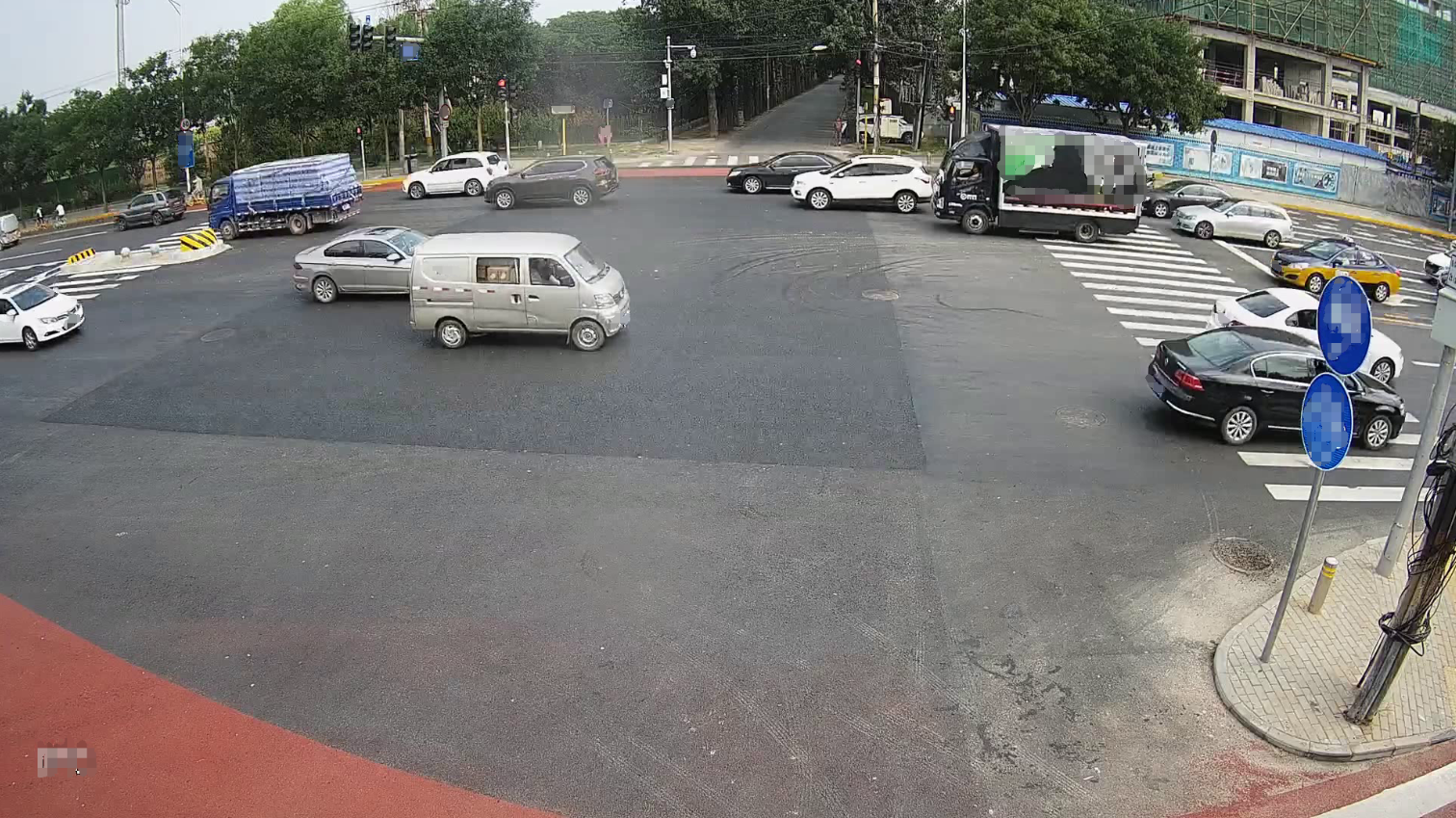} \\ \vspace{0.05in}
        \includegraphics[width=0.815\linewidth]{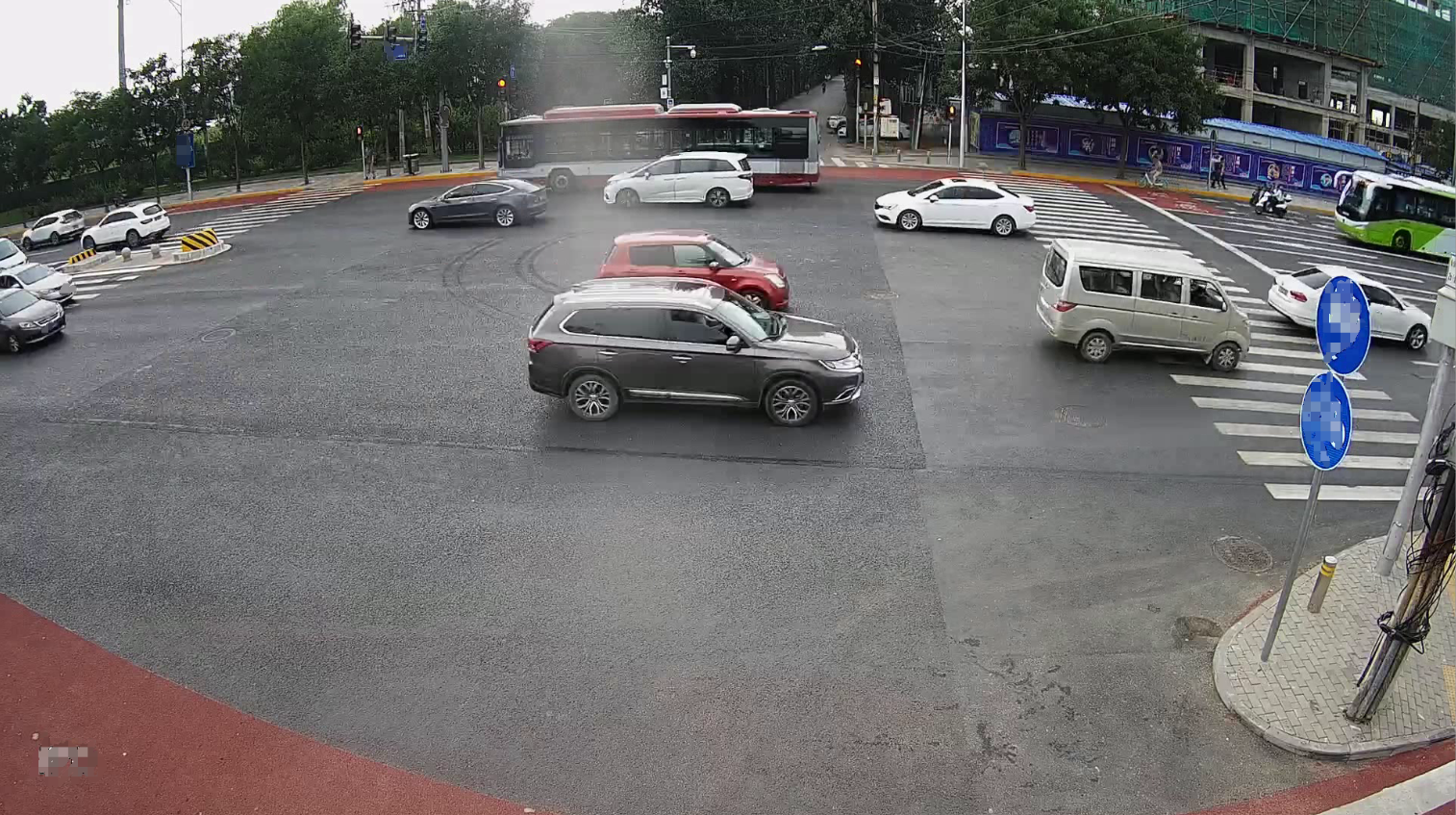} \\ \vspace{0.05in}
        \includegraphics[width=0.815\linewidth]{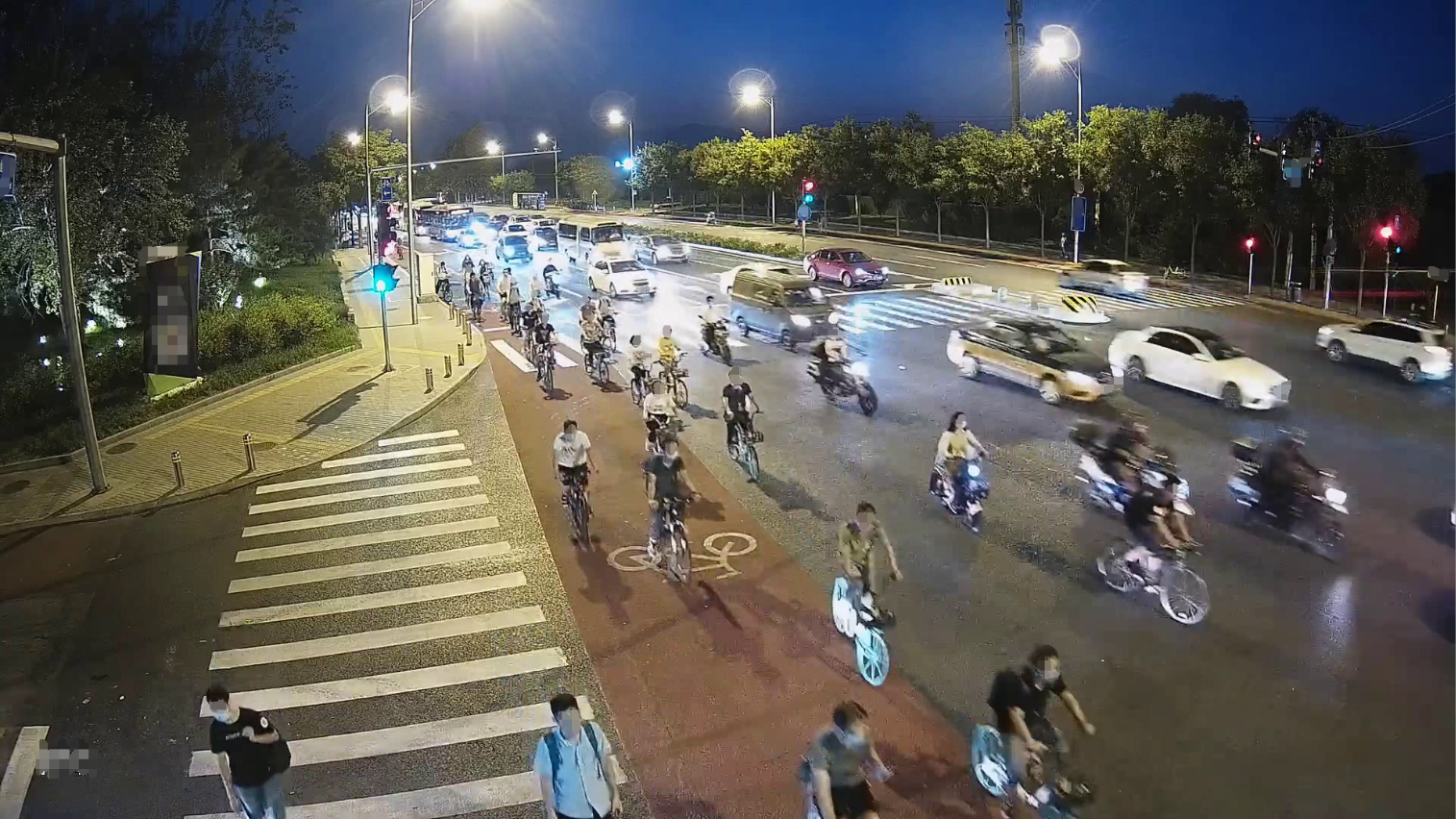} \\ \vspace{0.05in}
        \includegraphics[width=0.815\linewidth]{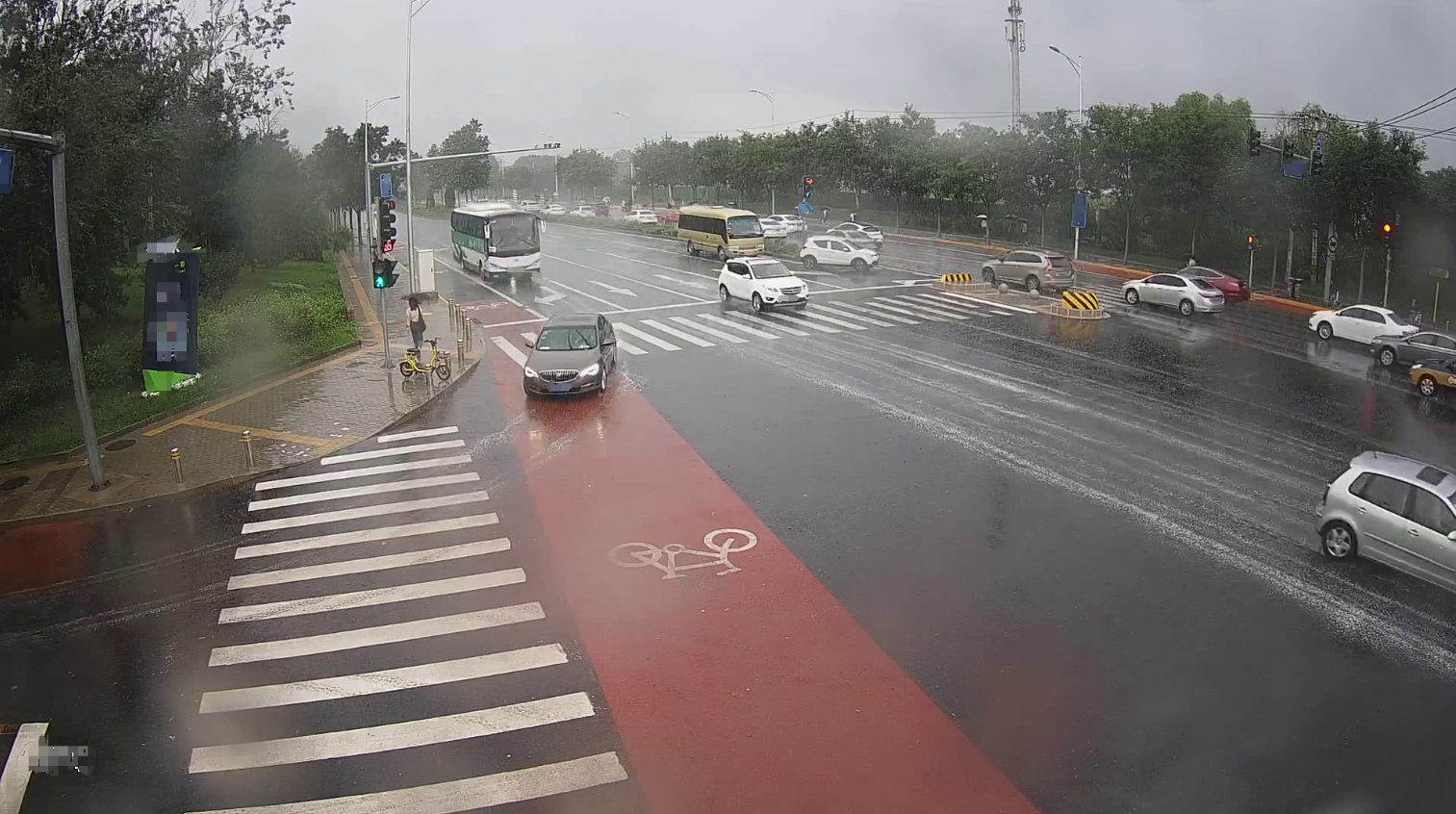} \\
    \end{center}
    \caption{A list of sample frames from the BAAI-VANJEE data collection.}
    \vspace{-0.2in}
    \label{fig:long}
\end{figure}

\subsection{Data Collection}

Our data were collected in urban intersections and highways by VANJEE smart base station placed on the roadside about 4.5m high in China.
The data were recorded under sunny, cloudy, and rainy weather conditions.
The VANJEE smart base station is equipped with a VANJEE 32L-LiDAR-R sensor(360° horizontal FOV) and four cameras(90° horizontal FOV) as showed in Figure 1.
By this way, we can greatly ensure the coverage and diversity of the collected data.
Figure 2 shows a list of sample LiDAR and RGB image frame unlabeled.

\subsection{Data Annotation}
To ensure the labeling quality, our annotations were labeled completely manually.
Owing to huge amount of objects in the annotation scene, it is time-consuming and costly to perform detailed annotations.
As far away or heavily occluded objects are easy to miss, our data annotation engineers paid particular attention to frames with dense object instances, ensuring there was no missing object annotation in such scenarios.

We collected more than one thousand hours of videos.
Some of the videos seemed not to be proper or useful, due to data quality, diversity, or other issues, for research and application purposes.
Therefore, we applied detailed and tailored data selection in terms of video quality, scenario diversity, and information security.
In BAAI-VANJEE dataset, we provided annotations of 12 classes of road objects. Similar to ApolloScape and {D$^2$-City}, tricycles were focused on as they often appear on the roads in China.
Refer to the corresponding camera FoV, we marked 3D bounding boxes, object classes, the heading angle directly on the LiDAR point cloud and 2D bounding boxes, object classes on the RGB images.
An illustration of our dataset is shown in Figure 3.
\begin{figure}[!htbp]
    \begin{center}
        \includegraphics[width=0.815\linewidth]{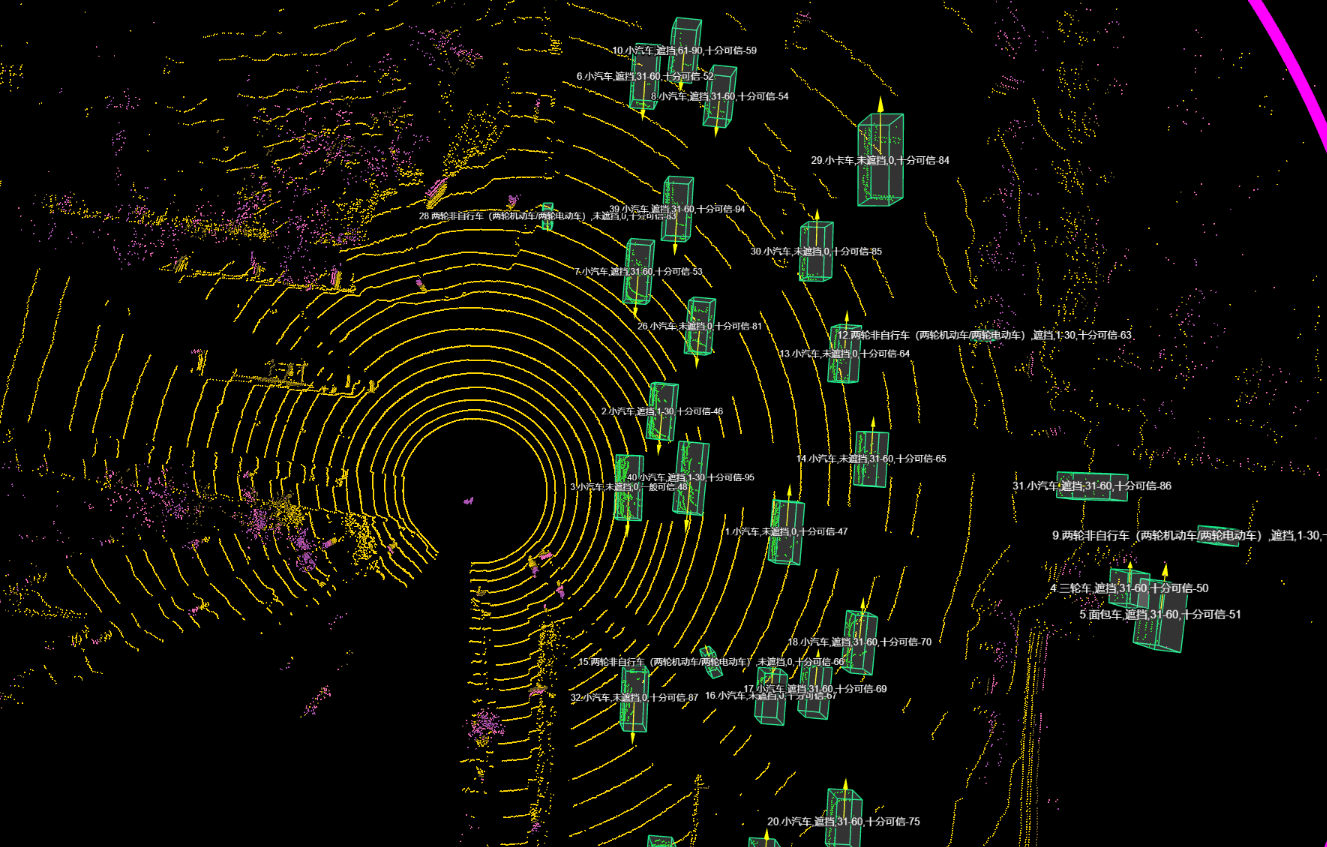} \\ \vspace{0.05in}
        \includegraphics[width=0.815\linewidth]{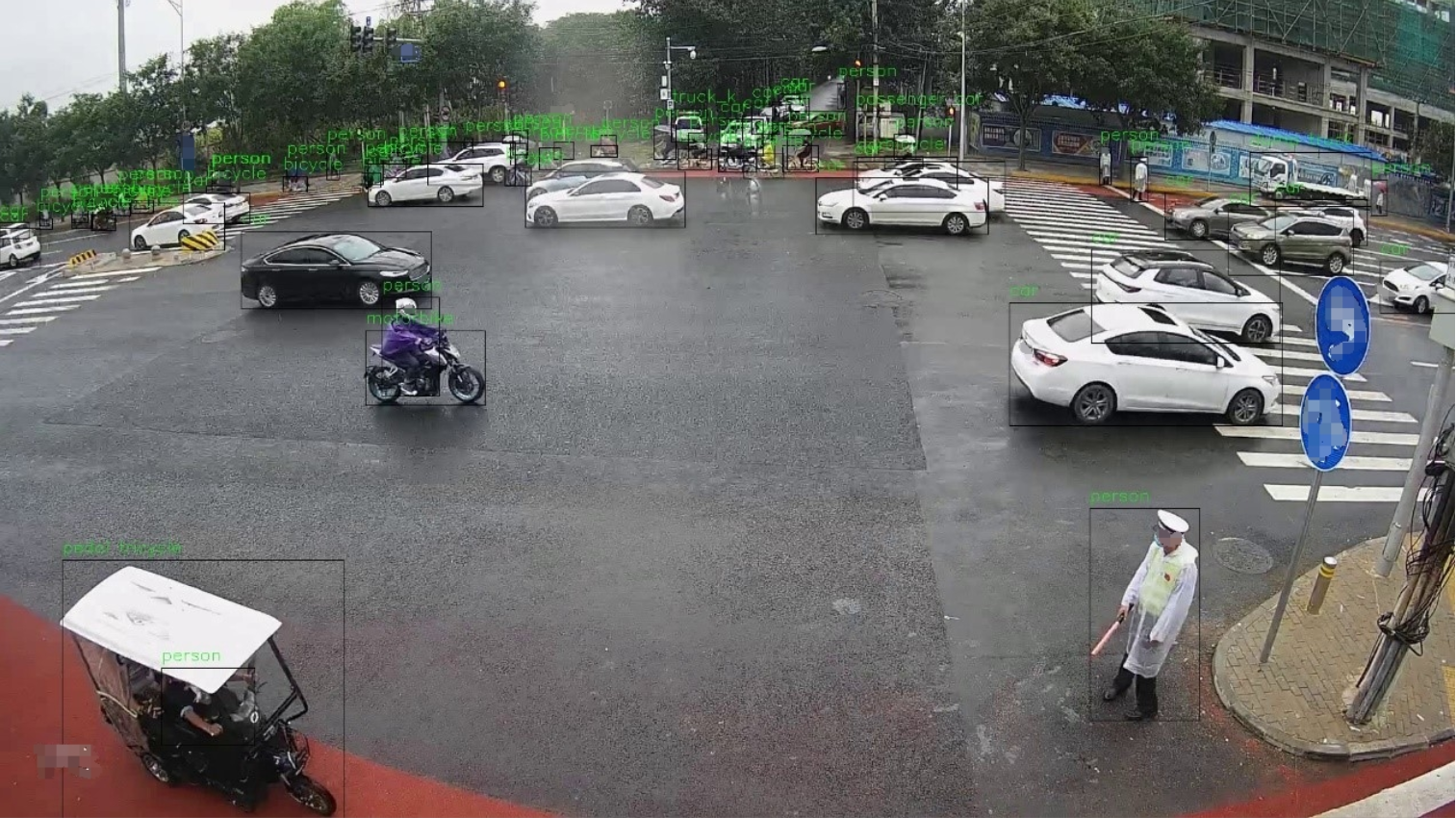} \\
    \end{center}
    \caption{Annotation samples from the proposed BAAI-VANJEE dataset with LiDAR data(Top) and RGB image(Bottom).}
    \vspace{-0.2in}
    \label{fig:long}
\end{figure}

\subsection{Label Statistics}
The BAAI-VANJEE dataset comprises 12 annotated classes corresponding to the most common objects in road scenes, named as pedestrian, bicycle, motorcycle, tricycle, car, van, cargo, truck, bus, semi-trailing tractor, special vehicle(police car, ambulance, fire truck, tanker truck, etc.) and roadblock.
Figure 4 illustrates number of annotation per category for our dataset. It can be seen from Figure 4(Top) that the number of annotations per frame is much higher than the KITTI dataset.
Comparison of the number of object for BAAI-VANJEE and KITTI with different radial distance away from LiDAR in meter is shown in Figure 5.
Objects more than 20m away from LiDAR of the BAAI-VANJEE dataset has much larger than the KITTI dataset, which demonstrates the over-the-horizon advantages of roadside sensors.
As shown in Figure 6, 37.42\% of the annotated objects(including LiDAR and image) are sunny data, 11.53\% are cloudy data, 27.88\% are nighty data and the remaining data are rainy, which demonstrates the diversity of samples.
Moreover, Figure 7 presents a reasonable size distribution(height, width, and length) of the most popular car objects for LiDAR data.

\begin{figure}[!htbp]
    \begin{center}
        \includegraphics[width=0.47\linewidth]{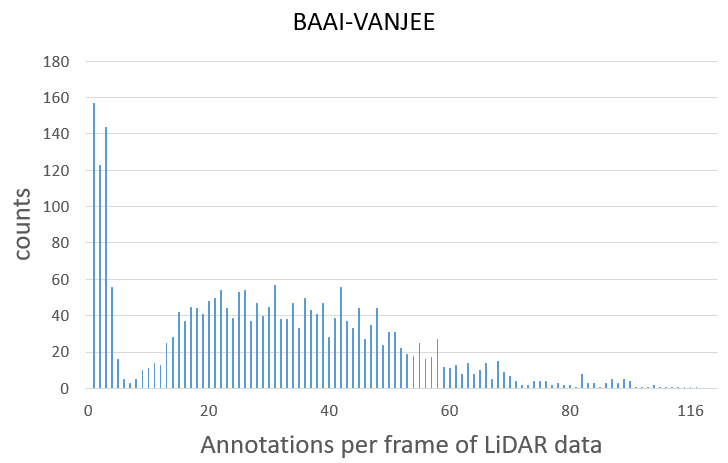} \hfill
        \includegraphics[width=0.47\linewidth]{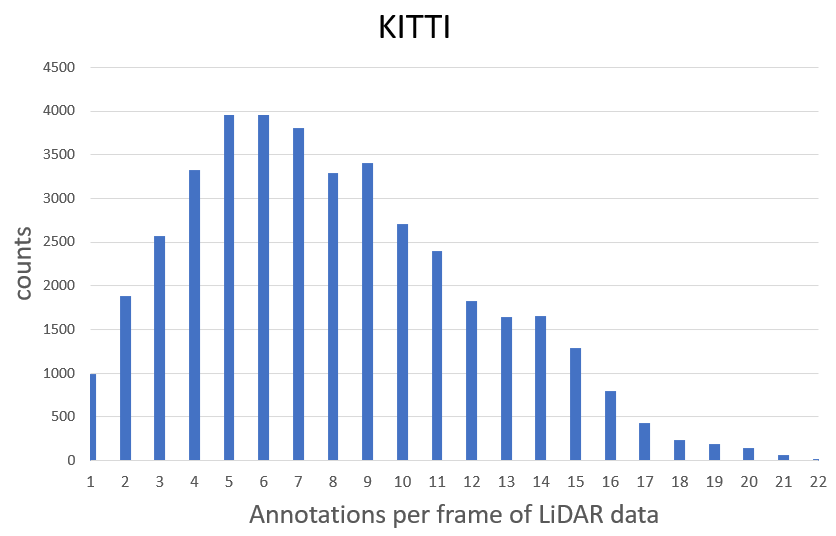} \hfill $\ $ \\ \vspace{0.05in}
        \includegraphics[width=0.95\linewidth]{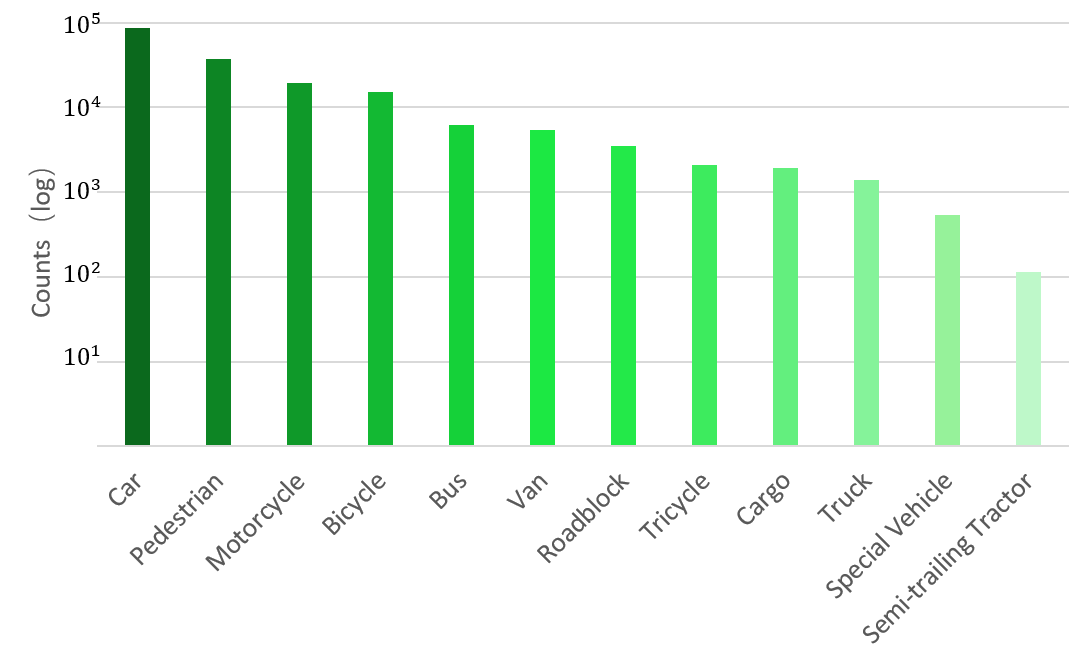} \hfill $\ $
    \end{center}
    \caption{Top: The number of object per frame for the BAAI-VANJEE dataset(left), and the KITTI dataset(right). Bottom: Number of annotation per category for the BAAI-VANJEE dataset.}
    \vspace{-0.2in}
    \label{fig:long}
\end{figure}
\begin{figure}[!htbp]
    \begin{center}
        \includegraphics[width=0.815\linewidth]{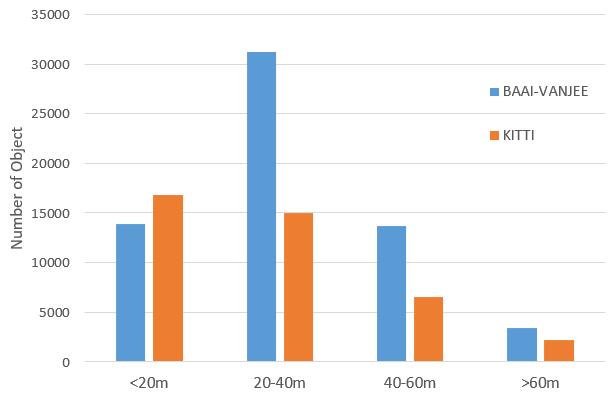} \\
    \end{center}
    \caption{Comparison of the number of object for BAAI-VANJEE and KITTI with different radial distance in meter.}
    \vspace{-0.2in}
    \label{fig:long}
\end{figure}
\begin{figure}[!htbp]
    \begin{center}
        \includegraphics[width=0.815\linewidth]{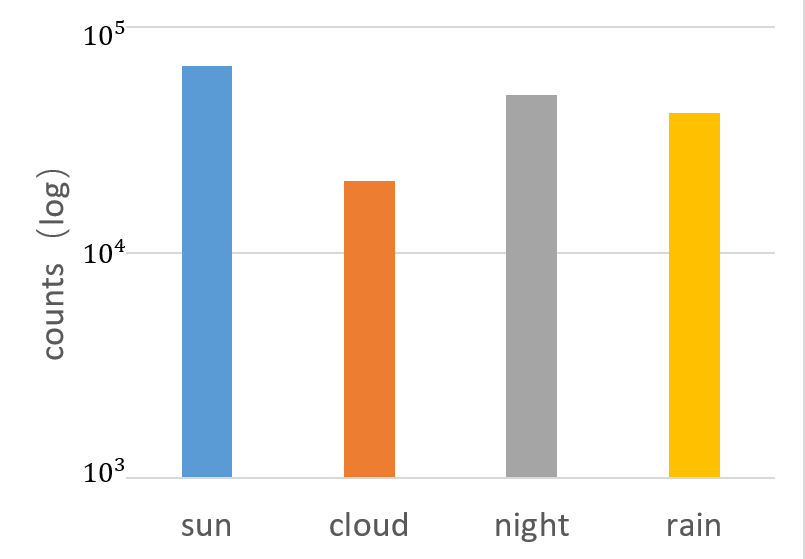} \\
    \end{center}
    \caption{Statistics on the numbers of annotated instances of different weathers.}
    \vspace{-0.2in}
    \label{fig:long}
\end{figure}
\begin{figure}[!htbp]
    \begin{center}
        \includegraphics[width=0.815\linewidth]{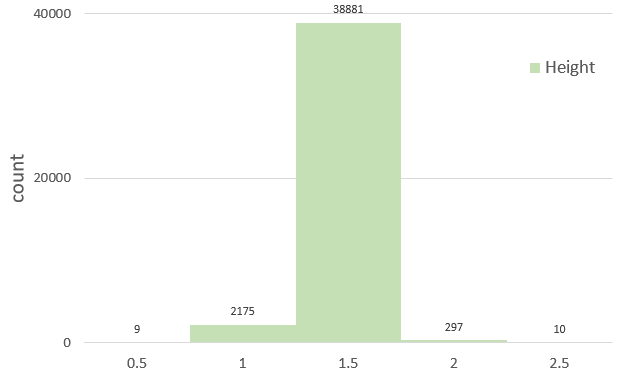} \\ \vspace{0.05in}
        \includegraphics[width=0.815\linewidth]{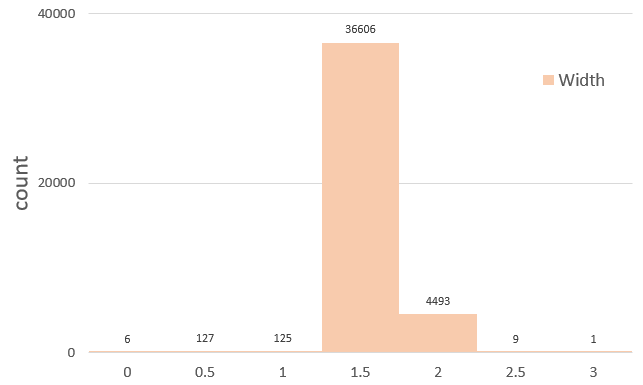} \\ \vspace{0.05in}
        \includegraphics[width=0.815\linewidth]{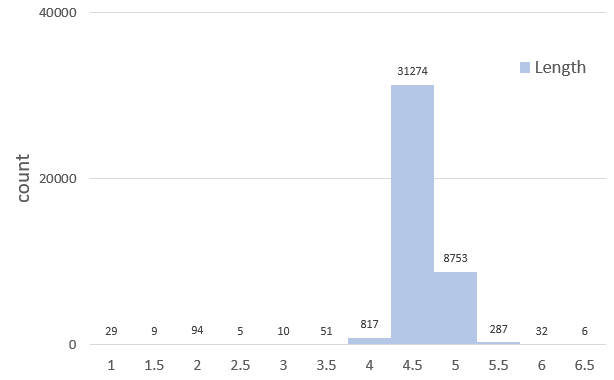} \\
    \end{center}
    \caption{Box dimensions(height [m], width [m], length [m]) for the car class.}
    \vspace{-0.2in}
    \label{fig:long}
\end{figure}

\section{Experiment: Object Detection}
In this section, we present the results of training and evaluating a 3D object detection method on BAAI-VANJEE dataset and COCO dataset~\cite{chen2015coco}.
For evaluating our experiment, we defined a standard validation set, which has various desirable properties including sun/rain, day/night frames.
In more details, we carefully selected $1000$ LiDAR frames from the BAAI-VANJEE dataset as the validation set.
We selected widely used 3D object detector that are publicly available online, namely SA-SSD.
For the datasets to be evaluated, we used the same pre-training weights.
We trained the network using COCO dataset and BAAI-VANJEE dataset respectively.
The 3D object detection performance was investigated on the most common car, bicycle and pedestrian in autonomous driving with the validation set.
The evaluation metric is Average Precision(AP) at 0.25 IoU.

Evaluation results are shown in Table 1.
It is clearly that the model with BAAI-VANJEE dataset achieves a better result.
Due to the lack of public roadside datasets, we selected BAAI-VANJEE dataset without training as the validation set.
Therefore, BAAI-VANJEE dataset is more friendly to roadside perspective perception.
\begin{table}[t]
    \begin{center}
        \begin{tabular}{c|c|c|c}
            \toprule
            & \textit{car} & \textit{bicycle} & \textit{pedestrian} \\ \cmidrule{1-4}
            COCO & $64.54\%$ & $13.42\%$ & $4.55\%$ \\ \cmidrule{1-4}
            BAAI-VANJEE & $88.62\%$ & $76.12\%$ & $63\%$ \\
            \bottomrule
        \end{tabular}
    \end{center}
    \caption{Results of COCO and BAAI-VANJEE datasets on 3D Car, Bicycle and Pedestrian detection(AP).}
    \label{tab:obj-area}
\end{table}

\section{Conclusion and Future Work}
In this work, we propose a challenging BAAI-VANJEE roadside dataset(LiDAR data and RGB images) for detection tasks for CAVH in challenging environments, which contains $2500$ frames of LiDAR data, $5000$ frames of RGB images and detailed annotations of all frames.
As the first public roadside dataset, BAAI-VANJEE dataset aims to actively assist the academic and industrial circles to accelerate the innovation research and achievement transformation in the field of intelligent transportation in big data era.
In the future, we plan to release more diverse and complex roadside datasets, especially the tracking annotated dataset.

{\small
\bibliographystyle{ieee}
\bibliography{reference}
}

\appendix
\section{Class Examples}
Figure 8 shows some object examples of each class to be annotated in the LiDAR and image frames.
\begin{figure*}[htb]
    \begin{center}
        \includegraphics[width=0.815\linewidth]{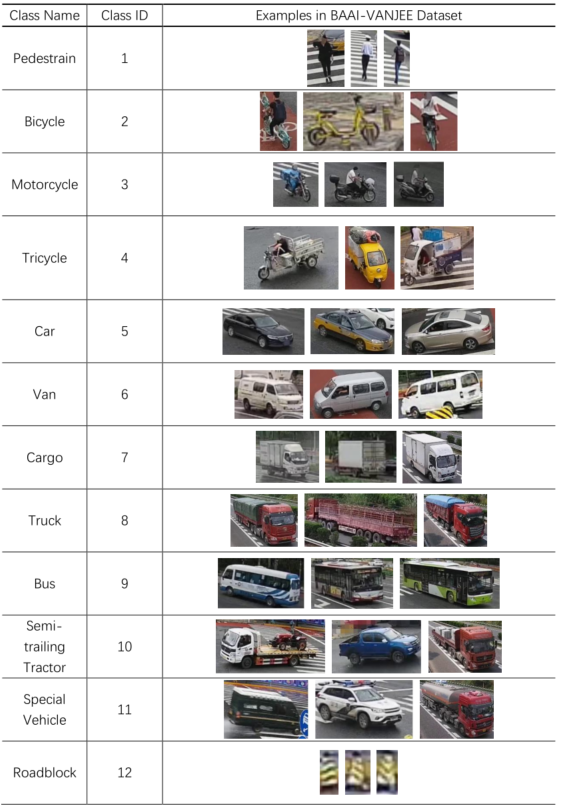}\\
    \end{center}
    \caption{Examples of the 12 annotated classes of objects.}
    \label{fig:obj-area}
\end{figure*}

\end{document}